\pdfoutput=1

\documentclass[11pt]{article}

\usepackage{acl}

\usepackage{times}
\usepackage{latexsym}

\usepackage{nert}

\usepackage{tikz}
\usetikzlibrary{trees}

\usepackage[T1]{fontenc}

\usepackage{lipsum}

\usepackage{microtype}

\title{Measuring Fine-Grained Semantic Equivalence\\with Abstract Meaning Representation}

\author{Shira Wein \\
  Georgetown University \\
  \textsmaller[.5]{\eml{sw1158@georgetown.edu}} \\\And
  Zhuxin Wang \\
   Georgetown University \\
  \textsmaller[.5]{\eml{zw85@georgetown.edu}} \\\And
  Nathan Schneider \\
  Georgetown University \\
  \textsmaller[.5]{\eml{nathan.schneider@georgetown.edu}}}

\begin{document}
\maketitle
\begin{abstract}

Identifying semantically equivalent sentences is important for many cross-lingual and monolingual NLP tasks.
Current approaches to semantic equivalence take a loose, sentence-level approach to ``equivalence,'' despite previous evidence that fine-grained differences and implicit content have an effect on human understanding \citep{roth-anthonio-2021-unimplicit} and system performance \citep{briakou-carpuat-2021-beyond}. 
In this work, we introduce a novel, more sensitive method of characterizing semantic equivalence that leverages Abstract Meaning Representation graph structures. 
We develop an approach, which can be used with either gold or automatic AMR annotations, and demonstrate that our solution is in fact finer-grained than existing corpus filtering methods and more accurate at predicting strictly equivalent sentences than existing semantic similarity metrics. 
We suggest that our finer-grained measure of semantic equivalence could limit the workload in the task of human post-edited machine translation and in human evaluation of sentence similarity. 



\end{abstract}

\section{Introduction}

Translation between two languages is not always completely meaning-preserving, and information can be captured by one sentence which is not captured by the other.
For example, consider the parallel French and English sentences from the REFreSD dataset (an annotated subset of the French-English WikiMatrix \citep{schwenk2019wikimatrix}) shown in \cref{fig:religious_implicit_example}.
The French sentence says ``tous les autres édifices'' (\emph{all other buildings}) while the English specifies ``all other \emph{religious} buildings.'' Because the sentence goes on to list  religious buildings, it could be inferred from context that the French is describing other \emph{religious} buildings. 
The French author, for whatever reason, chose to exclude \emph{religious}; the sentences thus convey the same overall meaning but are not \emph{exactly} parallel.

Semantic divergence (or conversely, semantic equivalence) detection aims to pick out parallel texts which have less than equivalent meaning.
Under a strict or close analysis of the translation, these sentences could be considered divergent, because the meanings are not identical (in particular, information made explicit in one is left implicit in the other).
But in the REFreSD corpus \cite{briakou-carpuat-2020-detecting}, the sentences are annotated as having no meaning divergence
because at the sentence-level they are essentially equivalent.

Semantic divergence detection plays an important role in many cross-lingual NLP tasks, such as text reuse detection, translation studies, and filtering of parallel corpora or MT output.
Though semantic divergence across sentences in parallel corpora has been well-studied,
current detection methods fail to capture the full scope of semantic divergence.
State-of-the-art semantic divergence systems rely on perceived \emph{sentence-level divergences}, which do not entirely encapsulate all semantic divergences. Two sentences are considered equivalent (non-divergent) at the sentence level if the same overall information is conveyed, even if there are minor meaning differences.
Finer-grained differences are not widely considered in the detection of semantic divergences, despite the fact that implicit information can be critical to the understanding of the sentence \cite{roth-anthonio-2021-unimplicit}.


\begin{figure}[t]
\small
All other \emph{religious} buildings are mosques or Koranic schools founded after the abandonment of Old Ksar in 1957.\\[5pt]
Tous les autres édifices sont des mosquées ou des écoles coraniques fondées à l'époque postérieure à l'abondance du vieux ksar en 1957.
\caption{Two parallel sentences from the REFreSD dataset marked as having no meaning divergence, for which the AMRs diverge. 
 }
  \label{fig:religious_implicit_example}
\end{figure}


We argue that a finer-grained measure of semantic equivalence is needed: a way to detect \emph{strictly} semantically equivalent sentence pairs.
In this work, we develop an approach to \emph{strict} semantic equivalence detection. 
Our approach to the identification of semantic equivalence moves beyond surface-level perception of divergence and accounts for more subtle differences in meaning that cannot be detected by simply comparing the words of the sentences.
We leverage the Abstract Meaning Representation  \citep[AMR;][]{banarescu-etal-2013} formalism to identify these divergences at a fine-grained level.
We present an approach to semantic equivalence detection via AMR, with analysis of data in two language pairs (English-French and English-Spanish).
Drawing on previous work on semantic divergence (\cref{sec:background}),
we demonstrate that sentence-level divergence annotations can be coarse-grained, neglecting slight differences in meaning (\cref{sec:amr_for_divergence}). 
%
We find that comparing two AMR graphs is an effective way to characterize meaning in order to uncover finer-grained divergences (\cref{sec:data}),
%
and this can be achieved even with automatic AMR parsers 
(\cref{sec:automatic_parses}).
Finally, in \cref{sec:bertscore} we evaluate our AMR-based metric on a cross-linguistic semantic textual similarity dataset, and show that for detecting semantic equivalence, it is more precise than a popular existing model, multilingual BERTScore \citep{zhang2019bertscore}.

Our primary contributions include:
\begin{itemize}
\item Our novel approach to the identification of semantic divergence---which moves beyond perceived sentence-level divergences---through the lens of AMR
\item A simple pipeline algorithm that modifies Smatch to quantify divergence in cross-lingual AMR pairs, automating the detection of AMR-level divergence
\item Studies demonstrating that our AMR-based approach accurately captures a finer-grained degree of semantic equivalence than both the state-of-the-art corpus filtering method and a semantic textual metric
\end{itemize}
We will release the code and dataset for this work upon publication to enable the use of AMR for semantic divergence detection.




\section{Background on Semantic Divergence}
\label{sec:background}


Semantic divergences can arise when translating from one language or another. These divergences can arise due to different language structure, syntactic differences in the language, or translation choices \cite{dorr-1994-machine,doorsolvingthematicdivergences}. Additional divergences can be introduced when automatically extracting and aligning parallel resources \cite{smith-etal-2010-extracting,zhai-etal-2018-construction,fung-cheung-2004-multi}.

To address these divergences, a number of  systems have been developed to automatically identify divergences in parallel texts \cite{carpuat-etal-2017-detecting,vyas-etal-2018-identifying,briakou-carpuat-2020-detecting,briakou-carpuat-2021-beyond,zhai-etal-2020-detecting}.
The approach taken by \citet{briakou-carpuat-2020-detecting} to detecting sentence-level semantic divergences involves training multilingual BERT \cite{DBLP:journals/corr/abs-1810-04805} to rank sentences diverging to various degrees. They introduced and evaluated on a novel dataset called Rational English-French Semantic Divergences (REFreSD). REFreSD consists of English-French parallel sentences, with crowdsourced annotations classifying the sentences as having no meaning divergence, having some meaning divergence, or being unrelated.



Recent work has investigated the differences in cross-lingual (English-Spanish) AMR pairs within the framework of translation divergences \citep{wein-schneider-2021-classifying}. Specifically, this work developed an annotation schema to classify the types and causes of differences between cross-lingual AMR pairs.
We use this dataset to test the performance of our system on English-Spanish gold AMR pairs. (For English-French, we produce our own gold judgments of AMR divergence to test our algorithm.)
Additional prior work has explored the role of structural divergences in cross-lingual AMR parsing \citep{blloshmi-etal-2020-xl,damonte_thesis}.

\section{AMR for Identification of Semantic Equivalence}
\label{sec:amr_for_divergence}

Semantic representations are designed to capture and formalize the meaning of a sentence.
%
In particular, the Abstract Meaning Representation (AMR) framework aims to formalize sentence meaning as a graph in a way that is conducive to broad-coverage manual annotation \citep{banarescu-etal-2013,AMRGuidelines}. These semantic graphs are rooted and labeled, such that each node of the graph corresponds to a semantic unit. AMR does not capture nominal or verbal morphology or many function words,
 abstracting away from the syntactic features of the sentence.
Attributes are labeled on edges between nodes (concepts), and these attributes can be either core roles / arguments, marked with \texttt{:ARG0}, \texttt{:ARG1}, etc., or non-core roles such as \texttt{:time}, \texttt{:domain}, and \texttt{:manner}. The root of the AMR is usually the main predicate of a sentence.


\begin{figure}[tb]
\small

He later scouted in Europe for the Montreal Canadiens.
\begin{verbatim}
(s / scout-02
      :ARG0 (h / he)
      :ARG1 (c / continent
            :wiki "Europe"
            :name "Europe")
      :ARG2 (c2 / canadiens
            :mod "Montreal")
      :time (a / after))
\end{verbatim}

\smallbreak
Il a plus tard été dépisteur du Canadiens de Montréal en Europe. (\emph{He later scouted for the Montreal Canadiens in Europe.})
\begin{verbatim}
(d / dépister-02
      :ARG0 (i / il)
      :ARG1 (c / continent
            :wiki "Europe"
            :name "Europe")
      :ARG2 (c2 / canadiens
            :mod "Montreal")
      :time (p / plus-tard))
\end{verbatim}
  \caption{A pair of sentences and their human annotated AMRs, for which the sentences receive a ``no meaning divergence'' judgment in the REFreSD dataset, and are also equivalent per AMR divergence.}
  \label{fig:parallel_scout_example}
\end{figure}



We leverage the semantic information captured by AMR to recognize semantic equivalence or divergence across parallel sentences. 
\Cref{fig:parallel_scout_example}, for example, illustrates a strictly meaning-equivalent sentence pair along with the AMRs. 
Though the sentences differ with respect to syntax and lexicalization, the AMR graphs are structurally isomorphic.
If the AMR structures were to differ, that would signal a difference in meaning.

Two particularly beneficial features of the AMR framework are the rooted structure of each graph, which elucidates the semantic focus of the sentence, as well as the concrete set of specific non-core roles, which are useful in classifying the specific relation between concepts\slash semantic units in the sentence.
For example, in \cref{fig:amr_divergence}, the emphasis on the English sentence is on possession---\emph{your} planet---but the emphasis on the Spanish sentence is on place of origin, asking, which planet are you \emph{from?} This difference in meaning is reflected in the diverging roots of the AMRs.

We find that non-core roles (such as \texttt{:manner}, \texttt{:degree}, and \texttt{:time}) are particularly helpful in identifying parallelism or lack of parallelism between the sentences during the annotation process. This is because AMR abstracts away from the syntax (so that word order and part of speech choices do not affect equivalence), but instead explicitly codes relationships between concepts via semantic roles.
Furthermore, AMRs use special frames for certain relations, such as \texttt{have-rel-role-91} and \texttt{include-91}, which can be useful in enforcing parallelism when the meaning is the same but the specific token is not the same. For example, if the English and French both have a concession, but the English marks it with ``although'' and the French marks it with ``\emph{mais}'' (but), the special frame role will indicate this concession in the same way, preserving parallelism. 

\paragraph{Granularity of the REFreSD dataset.} 
Another example, using sentences from the REFreSD dataset, is shown in  \cref{fig:cd_sales_example}. These sentences are marked as having no meaning divergence in the REFreSD dataset but do have diverging AMR pairs. The difference highlighted by the AMR pairs is the \texttt{:time} role of reach / \emph{atteindre}. The English sentence says that no.~1 is reached ``within a few weeks'' of the release, while the French sentence says that no.~1 is reached the first week of the release (\emph{la première semaine}). \textbf{In examples like this one it is made evident that sentence-level divergence (as appears in REFreSD) do not capture all meaning differences.}

\begin{figure}[tb]
\small

Which is your planet?
\begin{verbatim}
(p / planet
    :poss (y / you)
    :domain (a / amr-unknown))
\end{verbatim}

\smallbreak
¿ De qué planeta eres ? (\emph{Which planet are you from?})
\begin{verbatim}
(s / ser-de-91
   :ARG1 (t / tú)
   :ARG2 (p / planeta
       :domain (a / amr-desconocido)))
\end{verbatim}
  \caption{Two parallel sentences and AMRs from the \citeauthor{migueles-abraira-etal-2018-annotating} English-Spanish AMR dataset, which diverge in meaning. The Spanish role labels are translated  into English here for ease of comparison.}
  \label{fig:amr_divergence}
\end{figure}

\begin{figure}[tb]
\bigbreak
\small
Although the sales were slow (admittedly, according to the band), the second single from the album, "Sweetest Surprise" reached No. 1 in Thailand \emph{within a few weeks} of release. \\[5pt]
%
Même si les exemplaires ont du mal à partir (comme l'admet le groupe), le second single de l'album, Sweetest Surprise, atteint la première place en Thaïlande \emph{la première semaine} de sa sortie.
\caption{Two parallel sentences from the REFreSD dataset \cite{briakou-carpuat-2020-detecting} marked as having no meaning divergence, but for which the AMRs diverge. Italicized spans indicate the cause of the AMR divergence.
 }
  \label{fig:cd_sales_example}
\end{figure}

We explore the ability to discover semantic divergences in sentences either with gold parallel AMR annotations or with automatically parsed AMRs using a multilingual AMR parser, in order to enable the use of this approach on large corpora (considering that AMR annotation requires training).

We propose that an approach to detecting divergences using AMR will be a stricter, finer-grained measurement of semantic divergence than perceived sentence-level judgments. The use of a finer-grained metric would enable more effective filtering of parallel corpora to sentences which have minimal semantic divergence.



\section{Examining and Automatically Detecting Differences in Gold AMRs}
\label{sec:data}

In this section, we \textbf{evaluate the ability of AMR to expose fine-grained differences in parallel sentences} and how to \textbf{automatically detect those differences}.
In order to do so, we produce and examine English-French AMR pairs, which is the first annotated dataset of French AMRs.

\subsection{Examination of Gold AMR Data}
\label{ssec:examination}

We focus on French for effective comparison with sentence-level semantic divergence models (because of the available resources), though it also makes for ideal candidates in a cross-lingual AMR comparison, as it is broadly syntactically similar to English. This suggests that the AMRs could be expected to look similar (though not exactly the same) as inflectional morphology and function words are not represented in AMR. Prior work has investigated the transferability of AMR to languages other than English, and has found that it is not exactly an interlingua, but in some cases cross-lingual AMRs align well. Additionally, some languages are more compatible (Chinese) with English AMR than other languages (Czech) \citep{xue-etal-2014-interlingua}.


\paragraph{English-French AMR Parallel Corpus}

In investigating the differences between the degree of divergence captured by AMR and sentence-level divergence, we aim to compare quantitative measures of AMR similarity with corresponding sentence-level judgments of similarity.
In order to compare human judgments and AMR judgments, we develop the first French-English AMR parallel corpus, which represents the first application of AMR to French. We produce gold AMR annotations for 100 sentences, which were randomly sampled, from the REFreSD dataset \citep{briakou-carpuat-2020-detecting,linh-nguyen-2019-case}.\footnote{We also test our system on the full REFreSD dataset, using an automatic AMR parser (described in \cref{sec:automatic_parses}).}
For the French AMR annotation process,
the role/argument labels were added in English as has been done in related non-English AMR corpora \citep{sobrevilla-cabezudo-pardo-2019-towards}, and the concept (node) labels were in French.\footnote{The specific concept sense numbers were based on English PropBank frames \citep{kingsbury-palmer-2002-treebank,palmer_2005_propbank}.}

\paragraph{Findings from Corpus Annotation}

\begin{table}[htb]
\centering
\small
\begin{tabular}{c | c | c} 
& AMR Div. & AMR Equi. \\
\midrule
Sentence-Level Div. & 57 & 0 \\
\midrule
Sentence-Level Equi. & 26 & 17 \\
\midrule
\end{tabular}
\caption{Comparison between AMR Divergence annotations and Sentence-Level Divergence REFreSD annotations for 100 French-English sentences. 
}
\label{tab:divergence_comparison}
\end{table}


In light of our research question considering whether AMR can serve as a proxy of fine-grained semantic divergence, we consider both qualitative and quantitative evidence.
While producing this small corpus of French-English parallel AMRs, our suspicions that AMR would be able to more fully capture semantic divergence than perceived sentence-level divergence were confirmed. We
uncovered a number of ways in which perceived sentence-level equivalence is challenged by the notion of AMR divergence.
Take the example in \cref{fig:religious_implicit_example}. The difference between ``religious'' being applied in the French sentence and appearing in the English sentence is not captured by perceived sentence-level divergence, but is captured by AMR divergence. 

Quantitative results appear in \cref{tab:divergence_comparison}. There are no instances where the sentence-level annotation claims that the sentences are divergent but the AMR annotations are equivalent. Conversely, there are 26 instances with AMR divergence but no perceived sentence-level semantic divergence.
From this annotation we find that AMR divergence is a finer-grained measure of divergence than perceived sentence-level divergence.

\label{ssec:spanish_corpus}

\subsection{Quantifying Divergence in Cross-Lingual AMR Pairs}
\label{ssec:algorithm}



We have shown that not all pairs that humans considered equivalent at the sentence level receive isomorphic AMRs because they actually contain low-level semantic divergences. This suggests AMRs can be useful for more sensitive automatic detection of divergence. 
Now, we investigate whether we can automatically detect and quantify this divergence on gold AMRs via the graph comparison algorithm Smatch. In order to quantify this divergence in cross-lingual AMR pairs, we develop a simple pipeline algorithm which is a modified version of Smatch and incorporates token alignment. We test our modified Smatch algorithm on gold English-French AMR pairs and gold English-Spanish AMR pairs in comparison to the similarity scores output by \citet{briakou-carpuat-2020-detecting}.
%

\paragraph{Modified cross-lingual version of Smatch.} 

Our simple pipeline algorithm extends Smatch, a measurement of similarity between two (English) AMRs \citep{cai-knight-2013-smatch}. 
Smatch quantifies the similarity of two AMRs by searching for an alignment of nodes between them that maximizes the $F_1$-score of matching (\textit{node1}, \textit{role}, \textit{node2}) and (\textit{node1}, \texttt{instance-of}, \textit{concept}) triples common between the graphs.
However, Smatch was designed to compare AMRs in the same language, with the same role and concept vocabularies.

To compare AMR nodes across languages, the nodes first need to be cross-lingually aligned. This involves translating the concept and role labels.
We take a simple approach of first word-aligning the sentence pair to ascertain corresponding concepts (most of which are lemmas of content words in the sentence).
Our approach is similar to that of \emph{AMRICA} \citep{saphra-lopez-2015-amrica}, but we use a different word aligner (fast\_align rather than GIZA++\footnote{fast\_align has been shown to produce more accurate word alignments,
such as in the case for Latvian-English translation \cite{latvianenglishgirgzdis}.}) and deterministic translation of role names if the labels are not in English. 
\finalversion{The deterministic translation is done using a mapping of the role names between Spanish and English provided in the Spanish annotation guidelines \citep{NoeliaMiguelesAbrairaMastersThesis}. 
The French role labels are already in English, so there is no need to translate them.\footnote{We release both versions of this code, for non-English AMRs with either English or non-English role labels.}}



\finalversion{We normalize the strings and remove sense labels from the English and French/Spanish concept labels.
An error that we noticed while developing the system was associated with the same concept label appearing more than once in either AMR, so we tag repeated words numerically before performing the alignment.}

\finalversion{Finally, we run Smatch with the default number of 4 random restarts to produce an alignment.}
The Smatch score produced is an F1 score from 0 to 1 where 1 indicates that the AMRs are equivalent. This can be converted to a binary judgment, where all non-1 pairs are divergent, or used as a continuous value (as in \cref{sec:automatic_parses}).
\finalversion{\nss{specify Smatch version}}




\paragraph{Testing our Approach on Gold AMRs.} One of the benefits of leveraging semantic representations in our approach to semantic divergence detection is that the identification of divergence boils down to determining whether the graphs are isomorphic or not (and accurate word alignment). This suggests that our pipeline algorithm (\cref{ssec:algorithm}) should be highly effective at identifying whether AMR pairs are divergent or equivalent. In order to test our AMR-based approach to strict semantic equivalence identification, we first test on gold AMRs, which are created by humans and thus have no external noise from being automatically parsed.

We expect that our AMR divergence characterization would behave differently from a classifier of sentence-level divergence. 
This is because the sentence-level classification methods require specialized training data and as such learn to classify based on the perceived sentence-level judgments of semantic divergence.
To test the strictness of our framing, we validate our quantification on gold English-French and gold English-Spanish cross-lingual AMR pairs.


\begin{table}[htb]
\centering
\small
\setlength{\tabcolsep}{5pt}
\begin{tabular}{c | c c c | c c c | c} 
\multicolumn{1}{c}{} & \multicolumn{3}{c}{\textbf{Equivalent} (17)} & \multicolumn{3}{c}{\textbf{Divergent} (83)} & \multicolumn{1}{c}{\textbf{All}} \\
\midrule
System & \textbf{P} & \textbf{R} & \textbf{F1} & \textbf{P} & \textbf{R} & \textbf{F1}  & \textbf{F1}  \\
\midrule
Ours & 1.00 & 0.82 & 0.90 & 0.97 & 1.00 & 0.98 & 0.97 \\
BC'20 & 0.39 & 0.82 & 0.53 & 0.95 & 0.73 & 0.83 & 0.75 \\
\end{tabular}
\caption{FR-EN: Binary divergence classification on 
on 100 gold French-English AMR pairs, annotated for sentences from the REFreSD dataset.
Precision (P), Recall (R), and F1 scores are reported for the
Equivalent, Divergent, and All AMR pairs. We compare the performance of our model with the performance of the \cite{briakou-carpuat-2020-detecting} model, referenced as BC'20, on our finer-grained measure of divergence for the same English-French parallel sentences.
}
\label{tab:french_english}
\end{table}

\paragraph{Results on Gold English-French AMR Pairs}
We test our pipeline algorithm on the 100 English-French annotated AMR pairs described in \cref{ssec:examination}. 
As expected, the simple pipeline algorithm is very accurate at correctly predicting whether the cross-lingual pairs do or do not diverge according to the stricter criterion.

\Cref{tab:french_english} showcases the ability of our pipeline system and the \cite{briakou-carpuat-2020-detecting} system (described in \cref{sec:background})
to identify these finer-grained semantic divergences. 
On these English-French AMR pairs, the F1 score for our system is 0.97 overall and 1.00 for equivalent AMR pairs. This high level of accuracy indicates we can reliably predict cross-lingual AMR divergence.

The \cite{briakou-carpuat-2020-detecting} system performs worse when using our finer-grained delineation of semantic divergence, achieving an F1 score of 0.75.\footnote{The \cite{briakou-carpuat-2020-detecting} system does not take AMRs as input, so we use the corresponding sentences as input for their system.}
Unsurprisingly, the precision, recall, and F1 for the their system is lower than the performance of our system, because theirs is not trained to pick up on these more subtle divergences. 
Note that on their own measure of divergence (perceived sentence-level divergence), the system achieves an F1 score of 0.85 on these same 100 sentences.

Of the 3 errors made by our algorithm (in all cases, classifying equivalent AMR pairs as divergent), 2 of the 3 are caused by word alignment errors. Named entities seem to pose an issue with fast\_align for our use case.

\begin{table}[htb]
\centering
\small
\setlength{\tabcolsep}{5pt}
\begin{tabular}{c | c c c | c c c | c} 
\multicolumn{1}{c}{} & \multicolumn{3}{c}{\textbf{Equivalent} (13)} & \multicolumn{3}{c}{\textbf{Divergent} (37)} & \multicolumn{1}{c}{\textbf{All}} \\
\midrule
System & \textbf{P} & \textbf{R} & \textbf{F1} & \textbf{P} & \textbf{R} & \textbf{F1}  & \textbf{F1}  \\
\midrule
Ours & 1.00 & 0.92 & 0.96 & 0.97 & 1.00 & 0.99 & 0.98 \\
BC'20 & 0.24 & 0.38 & 0.29 & 0.72 & 0.57 & 0.64 & 0.52 \\
\end{tabular}
\caption{EN-ES: Binary divergence classification with gold parallel AMRs. Included are Precision (P), Recall (R), and F1 for the Equivalent, Divergent, and All AMR pairs for our pipeline algorithm compared to the system by \citet{briakou-carpuat-2020-detecting}, referenced as BC'20, on the same English-Spanish parallel sentences.
} 
\label{tab:spanish_english}
\end{table}

\paragraph{Results on Gold English-Spanish AMR Pairs.}
In addition to testing our system on our English-French AMR annotations, we test our system on the 50 English-Spanish AMRs and sentences released by \citet{migueles-abraira-etal-2018-annotating}, who collected sentences from \emph{The Little Prince} and altered them to be more literal translations.
Recent work classified these AMRs via AMR structural divergence schema
\cite{wein-schneider-2021-classifying}
(mentioned in \cref{sec:background}).
\cite{wein-schneider-2021-classifying}

In \cref{tab:spanish_english}, we measure the ability of our pipeline system and the \cite{briakou-carpuat-2020-detecting} system 
to detect semantic divergences at a stricter level, as picked up by the AMR divergence schema.

Our system performs similarly well on Spanish-English pairs as it did on the English-French pairs, described in \cref{tab:french_english}.
This demonstrates that our pipeline algorithm is not limited to success on only one language pair, and we further affirm that the simple pipeline algorithm is a reliable way to predict cross-lingual AMR divergence.

\section{Strictness Results Using Automatic English-French AMR Parses}
\label{sec:automatic_parses}


In \cref{sec:data}, we showed that we are able to use gold (human annotated) AMRs to capture a finer-grained level of semantic divergence, quantifiable via Smatch.
We extend this further by determining whether fine-grained semantic divergences can be detected well even when using noisy automatically parsed AMRs.
To do so, we compare the the Smatch scores of automatically parsed AMR pairs with the human judgments output on the corresponding sentences by \citet{briakou-carpuat-2020-detecting}.

To take the expensive human annotation piece out of the process, we show that automatic AMR parses can be used instead of gold annotations by establishing a threshold, instead of via binary classification.
Therefore, we use the F1 score output by our pipeline algorithm as a \emph{continuous score} and establish thresholds to divide the data between divergent and equivalent.




We automatically parse cross-lingual AMRs for the entirety of the English-French parallel REFreSD dataset (1033 pairs).
The REFreSD dataset is parsed using the mbart-st version of SGL, a state-of-the-art multilingual AMR parser \cite{procopio-etal-2021-sgl}.
The (monolingual) Smatch score for the SGL parser, comparing our gold AMRs with the automatically parsed AMRs, is 0.41 for the 100 French sentences using Smatch (0.43 using our pipeline algorithm)\footnote{The SGL parser approaches cross-lingual parsing as the task of recovering the AMR graph for the English translation of the sentence, as defined in prior work \citep{damonte-cohen-2018-cross}. The result is that the parses of French sentences are largely in English, and default to French concepts only for out-of-vocabulary French words. The alignments in our pipeline account for this to better reward the native French concepts.} and 0.52 for the 100 parallel English sentences using Smatch.

In doing error analysis, we find that the data points which are classified as having no meaning divergence but have extremely low F1 scores are largely suffering from parser error. We do find that there are pairs classified in REFreSD as having no meaning divergence at the sentence-level that do correctly receive low F1 scores.
For example, the sentence pair in \cref{fig:cd_sales_example}, which has a REFreSD annotation of sentence-level equivalence and a gold AMR-level annotation of divergence, was assigned an F1 score of 0.3469.



Despite Smatch scores of 0.5 between the gold and automatic parses, both are usable for the task of detecting finer-grained semantic equivalence.
To demonstrate the usefulness of our continuous metric of semantic divergence using automatically parsed AMR pairs, we develop potential thresholds at which you could separate data as being equivalent vs.~divergent. 

\begin{figure}[htb]
    \centering 
    \includegraphics[trim={0cm 0cm 0.5cm 0.5cm},clip, scale=0.39]{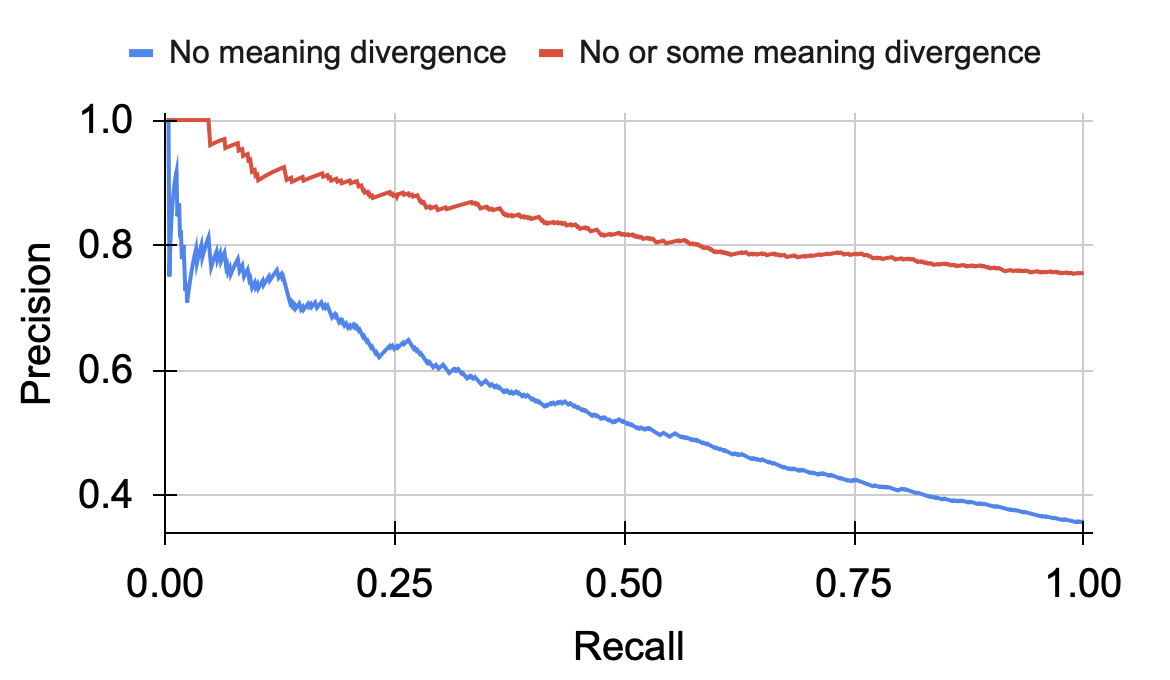}
    \caption{Precision / recall curve for equivalence detection in the 1033 sentence pairs in the full REFreSD dataset (English-French) using automatic AMR parses.
    Precision reflects the percent of sentences in which REFreSD human annotation was  equivalent (as labeled as no meaning divergence in the blue/bottom curve, or as labeled as having either no or some meaning divergence in the red/top curve).
    \finalversion{\nss{not urgent but something to consider for later: red horizontal line at 0.8 for majority baseline in the no-or-some setting. the red curve represents an inherently easier and less informative task} \sw{okay sounds good, I will leave it for now} }
    } 
    \label{fig:pc_curve}
\end{figure}

Because our metric is more sensitive, a practitioner could choose their own threshold by determining appropriate precision (how semantically equivalent they wanted a subset of filtered data to be) and recall (how much data they are willing to filter out) needs. This tradeoff is depicted in \cref{fig:pc_curve}. For example, if all pairs are marked as equivalent, precision would be approximately 40\% on the REFreSD dataset if considering solely the ``no meaning divergence'' pairs equivalent.



\paragraph{Comparing with model probabilities.}
Though it is reasonable to assume that if the gold AMR annotations provide a distinctly finer-grained measure of divergence than sentence-level divergence then this would also be the case when using automatically parsed AMRs, we want to ensure the continued strictness of our methodology. To do this, we compare the values of our continuous metric and the probabilities produced by the \cite{briakou-carpuat-2020-detecting} system. 

Because the probabilities produced by the system described in \cite{briakou-carpuat-2020-detecting} are always very close to 1 (equivalent) or very close to 0 (divergent) and there are far more divergent instances than equivalent instances, median and mode serve as a more effective form of comparison than mean between our F1 score and their probability score.
Above the 0.7 threshold, the median F1 for our system is 0.7869 and mode is 0.8; the median probability for the \citet{briakou-carpuat-2020-detecting} system is 0.9990 and the mode is 1.0.
For the 0.6 threshold, our median is 0.6667 and our mode is 0.6667; their median is 0.9871 and mode is 1.0.
Above the 0.5 threshold, our median is 0.5814 and our mode is 0.5; their median is 0.8907 and mode is 1.0.
Because these numbers are lower for our system than their system, we confirm that our measure is a stricter measure of equivalence even when using the automatically parsed AMRs.

If the goal is to prioritize items for a human to look at on a fixed budget, the absolute scores may matter less than rankings, though the rankings additionally differ drastically.
Of the top 50 sentences ranked by AMR divergence (which range in AMR similarity score from 0.96 to 0.67), only 19 of the 50 appear in the 166 sentences scored 1.0 by the \citet{briakou-carpuat-2020-detecting} system.


\section{Sentence Similarity Evaluation with Automatically Parsed English-Spanish AMRs}
\label{sec:bertscore}


Multilingual BERTscore \citep{zhang2019bertscore} is an embedding-based automatic evaluation metric of semantic textual similarity. Semantic textual similarity considers the question of semantic equivalence slightly differently because it rewards semantic overlap as opposed to equivalence.


As we have explored in previous sections, our AMR-focused approach in general is stricter than sentence-based measures of equivalence, in particular corpus filtering methods.
Because our system is a stricter measure of semantic equivalence, it may be the case that our system can more precisely identify the most similar sentences than existing measures of sentence similarity.
In this section, we look at the most semantically equivalent sentences in the dataset (as judged by our approach and as judged by multilingual BERTscore) in comparison to their human judgments of equivalence.
Specifically, we aim to investigate: (1)~whether the average human similarity score for the most similar \texttt{n} sentences is higher when ranked by our AMR-based metric versus when ranked by BERTscore, and (2)~whether human judgments of sentence similarity for the most similar sentences are more correlated with our AMR-based metric than with BERTscore. 
We compare our AMR-based metric to multilingual BERTscore because it has been shown to work will in cross-lingual settings when comparing system output to a reference \citep{koto-etal-2021-evaluating}.

\paragraph{Data.}
To perform this comparison, we use the 301 human annotated Spanish-English test sentences from the news down of the SemEval task on semantic textual similarity \citep{agirre-etal-2016-semeval}. 

\subsection{Smatch with Cross-Lingual AMR parsing}

For our analysis, we use the Translate-then-Parse system \citep[T+P;][]{uhrig-etal-2021-translate}.
Providing the Spanish sentences as input, T+P translates them into English, 
and then runs an AMR parser\footnote{Via amrlib: \url{https://github.com/bjascob/amrlib}} on the English translation. Because the Spanish sentence was translated into English and \emph{then} parsed, this automatic parse can be compared against the automatic parse of the original English sentence with plain Smatch (no cross-lingual alignment added). 

As we have established in \cref{sec:automatic_parses}, the noise introduced by automatic parsers can be overcome in our approach.
We validate that the Smatch scores retrieved after using \citeposs{uhrig-etal-2021-translate} parser still bears some correlation with the Smatch scores on the aligned gold AMRs.\footnote{On the 50 Spanish-English sentences mentioned in \cref{sec:data},
the correlation between the Smatch scores (in comparison to the same gold AMRs) when using either the translation-then-parse method or the method of aligning concepts via fast\_align is 0.31. This can be interpreted as a weak correlation. We find that both methods (translating the sentence first, or our pipeline algorithm aligning concepts in AMRs of different languages) work sufficiently well to capture the amount of divergence between cross-lingual AMR pairs.}


\subsection{Sentence Similarity Results}

\begin{figure}[tb]
    \centering 
    \includegraphics[trim={0cm 0cm 0cm 0cm},clip, scale=0.3]{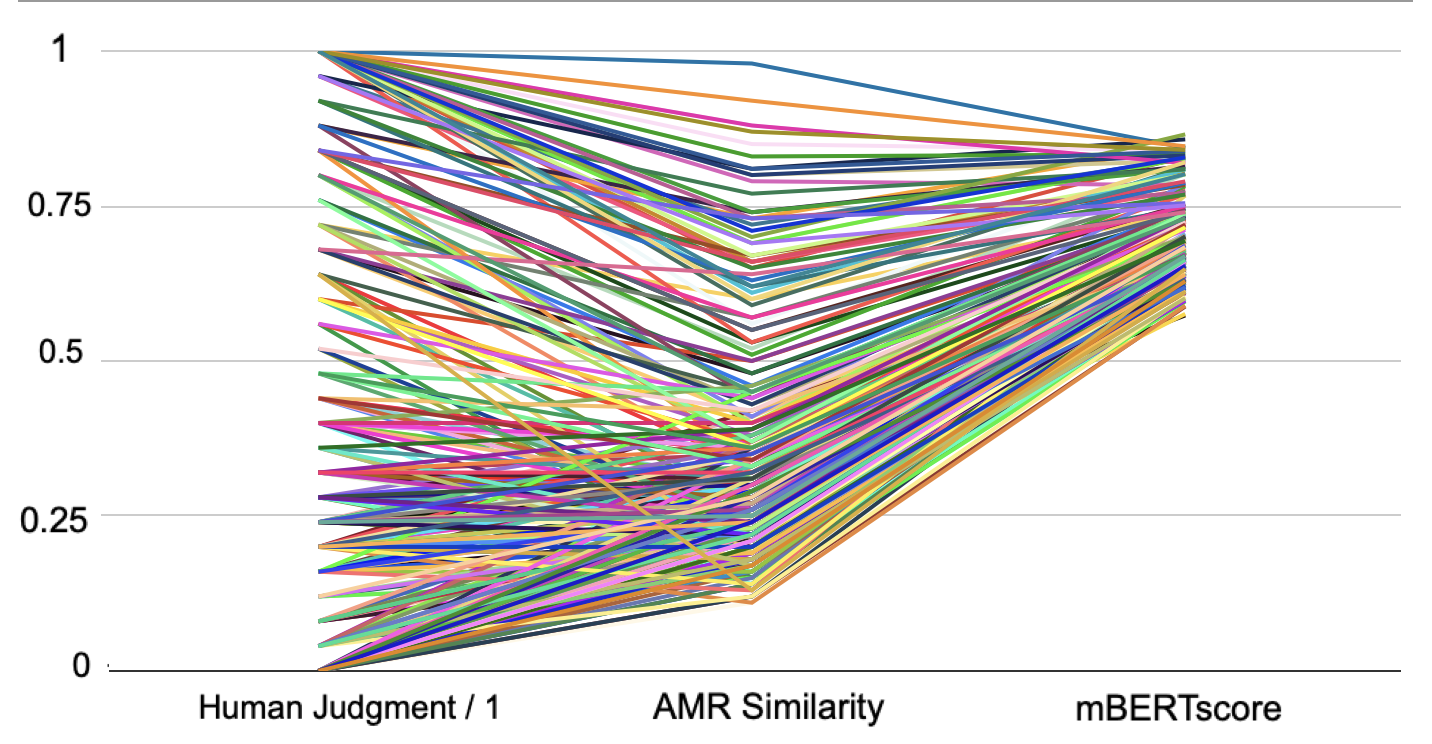}
    \caption{All data points normalized to a range of 0 to 1 for the Spanish-English sentence pairs from \citet{agirre-etal-2016-semeval}, including human judgment, AMR similarity score, and mBERTscore.
    } \label{fig:bertscore}
\end{figure}

The average human judgment score, on a scale of 0 to 5 with 5 being exactly equivalent, for all sentence pairs which have an AMR similarity score greater than 0.8 is 4.98. The average human judgment score for all sentence pairs which have an mBERTscore similarity score greater than 0.8 is 4.89. Similarly, 
the average human judgment score for pairs with an AMR similarity score of greater than 0.7 is 4.86, while the average human judgment score for pairs with an mBERTscore greater than 0.7 is 3.8. This is because mBERTscore takes a much broader view of semantic equivalence. While the human judgments occupy the full range of 0 to 5, the mBERTscores of these sentences range from 0.57 to 0.87, as shown in \cref{fig:bertscore}.
The AMR similarity score ranges from 0.11 to 0.98.

This might suggest that then a higher threshold should be used for mBERTscore to achieve the same level of semantic granularity.
However, our AMR similarity metric is also more correlated with human judgments for the most semantically equivalent sentences. For the top 20 items as ranked by AMR similarity, correlation with human judgments is 0.4068. But the top 20 items as ranked by mBERTScore are not correlated with human judgments ($-$0.0023).
When looking at all items above the mBERTscore of 0.8, correlation with human judgment is 0.1645, whereas for all items above the AMR similarity score of 0.8, correlation with human judgment is 0.2675. Correlation is calculated with Pearson correlation. Overall, AMR similarity score correlates with human judgment at a coefficient of 0.8367, which is slightly lower than the 0.8605 correlation between mBERTscore and human judgment.
This evidence further supports that our metric is in fact a finer-grained measure of semantic equivalence, and is therefore better at identifying which sentences are exactly semantically equivalent.





\section{Conclusion}

This paper introduced a new approach to finer-grained semantic equivalence detection, which is understudied but useful in a number of applications, including but not limited to reducing the workload of human translators in the task of post-editing machine translation.

We have effectively demonstrated that our definition of semantic divergence which leverages AMR is stricter than existing systems (\cref{ssec:algorithm}), when using either gold \emph{or} automatically parsed AMR pairs (\cref{sec:automatic_parses}). Further, we have shown that as a result our system is better at disambiguating similar sentences from strictly semantically equivalent sentences (\cref{sec:bertscore}).

We believe that our AMR-based framing of semantic divergence can be useful for decreasing the amount of data that needs to be post-edited by human translators or
annotated for human evaluation.
As it stands, MT systems which receive human post-editing, as well as translator aids which present MT output for human translators, present all sentences in the dataset to the human translator \citep{green2013efficacy}. Being able to filter out exactly semantically equivalent sentence pairs would reduce this workload. Similarly, filtering out exactly semantically equivalent sentences can lessen the amount of annotation necessary for human evaluations of text \citep{saldias2022toward}.

Other potential uses include cross-lingual text reuse detection (plagiarism detection), which asks whether one sentence is simply another sentence exactly translated \citep{potthast2011cross}. 
Translation studies and semantic analyses could also benefit from the distinction between semantically equivalent sentence pairs and sentence pairs which have subtle or implicit differences \citep{bassnett2013translation}.

\section{Acknowledgements}
Thank you to Rexhina Blloshmi, Eleftheria Briakou, and Luigi Procopio for providing system results, as well as to anonymous reviewers for their feedback. 
This work is supported by a Clare Boothe Luce Scholarship.

\bibliography{custom}
\bibliographystyle{acl_natbib}

\end{document}